# CIRO: COVID-19 infection risk ontology

**Shusaku Egami[1], Yasunori Yamamoto[2], Ikki Ohmukai[3], Takashi Okumura[4]***

**1** Artificial Intelligence Research Center, National Institute of Advanced Industrial Science and Technology (AIST), Koto, Tokyo, Japan, **2** Database Center for Life Science, Research Organization of Information and Systems, Kashiwa, Chiba, Japan, **3** Graduate School of Humanities and Sociology, The University of Tokyo, Bunkyo, Tokyo, Japan, **4** Health Administration Center, Kitami Institute of Technology, Kitami, Hokkaido, Japan

* taka@wide.ad.jp

**Data Availability Statement:** https://github.com/PLOD-info/PLOD/blob/master/rdf/CIRO.owl.

**Funding:** This work was supported in part by the Japan Agency for Medical Research and Development (AMED) under Grant JP20he0622042, and by Ministry of Health, Labour and Wel- fare (MHLW) under Program Grant Number 21HA2015. There was no additional external funding received for this study.

**Competing interests:** The authors have declared that no competing interests exist.

## Abstract

Public health authorities perform *contact tracing* for highly contagious agents to identify close contacts with the infected cases. However, during the pandemic caused by coronavirus disease 2019 (COVID-19), this operation was not employed in countries with high patient volumes. Meanwhile, the Japanese government conducted this operation, thereby contributing to the control of infections, at the cost of arduous manual labor by public health officials. To ease the burden of the officials, this study attempted to automate the assessment of each person's infection risk through an ontology, called COVID-19 Infection Risk Ontology (CIRO). This ontology expresses infection risks of COVID-19 formulated by the Japanese government, toward automated assessment of infection risks of individuals, using Resource Description Framework (RDF) and SPARQL (SPARQL Protocol and RDF Query Language) queries. For evaluation, we demonstrated that the knowledge graph built could infer the risks, formulated by the government. Moreover, we conducted reasoning experiments to analyze the computational efficiency. The experiments demonstrated usefulness of the knowledge processing, and identified issues left for deployment.

## 1 Introduction

Public health authorities perform contact tracing for highly contagious agents to identify and monitor the health status of persons in close contact with the infected cases. The tracing requires an in-depth survey of the cases, such as visiting places, means of transportation, persons contacted, and the way of interaction [1], to assess their infection risks for several days, or even weeks. In the pandemic caused by the new coronavirus disease 2019 (COVID-19), European countries and the United States did not perform contact tracing due to the large number of infection cases. However, in Japan, public health authorities conducted contact tracing, throughout the pandemic before the infections escalated. Isolation of contacts, coupled with other non-pharmaceutical interventions (NPIs) [2], contributed remarkably to patient reduction in Japan than in European countries. However, it was at the cost of the enormous manual labors made by public health officials.

If infection risks of individuals could be automatically estimated, identification of potential contacts would become quite efficient, making necessary arrangements for their testing and isolation more simple. Such a technology can be a game changer in NPIs to fight against contagious agents, such as measles, tuberculosis, and COVID-19. However, risk assessment of infections is a complex task that involves the processing of quantitative and qualitative information, as well as knowledge-based inference. Due to this complexity, contact tracing necessitated manual labor of public health officials, known as disease intervention specialists, and simple heuristics have been employed. In the COVID-19 pandemic, COVID-19 RECoVERY CONSULTING has proposed the "COVID-19 Risk Index" [3]. The Japanese government proposed "Three Cs" [4] that emphasize high-risk situations in the risk assessment: closed, crowded, and close-contact settings. Even with the heuristics, the estimation of the infection risk of individual cases necessitated processing of the knowledge on behavioral histories, spatial conditions of locations visited, personal risk factors, and others, all of which have made the automation impractical. The structured and standardized knowledge sources of the actions and the spaces are the keys to the solution.

Therefore, in this study, we propose an efficient mechanism that supports the contact tracing and screening of people who have had close contact, by applying semantic technologies to organize the quantitative and qualitative information associated with infections. First, we developed COVID-19 Infection Risk Ontology (CIRO) that can infer the infection risk in the action history. The CIRO can represent not only a patient's action and contact history, but the risks associated with the properties and situations of the space, risks caused by human behavior, and relationship with the actions induced by the space based on an event-centric model. Next, dummy data that imitate the actual behavior survey data are generated, and converted into a knowledge graph (KG) in Resource Description Framework (RDF) format [5], using the constructed ontology. On this KG, we demonstrate that SPARQL Protocol and RDF Query Language (SPARQL) [6], and Web Ontology Language (OWL) [7] reasoning can be used to search for risky events and related persons. Finally, for practical application in the contact tracing work, we evaluated the feasibility of the proposed method by measuring the validity and execution speed of the inference results while changing the scale of the simulated data.

The reminder of the paper is structured as follows: Section 2 introduces related studies. Section 3 describes the proposed CIRO. Section 4 evaluates the ontology by experiments to assess infection risks through searching and inference. Section 5 discusses the results from ontological and practical perspectives. Finally, Section 6 presents the conclusion. The experiments show that the ontology realized through KG technologies standardized by the World Wide Web Consortium (W3C) can be a promising tool in NPIs against pandemics by drastically reducing the manual labor required for contact tracing operations by public health authorities.

## 2 Related works

KG, the basis of this study, is "*a graph of data intended to accumulate and convey knowledge of the real world, whose nodes represent entities of interest and whose edges represent potentially different relations between these entities*", according to Hogan et al. [8]. However, it can be defined in several other ways. Ehrlinger et al. [9] defined a KG as something that "*acquires and integrates information into an ontology and applies a reasoner to derive new knowledge*." Feilmayr et al. [10] defined an ontology as "*a formal, explicit specification of a shared conceptualization that is characterized by high semantic expressiveness required for increased complexity*." According to previous reports [10], "As required in artificial intelligence, an ontological model allows known facts and/or assumptions to be used to derive a conclusion or to make inferences (i.e., reasoning)." In what follows, we refer to the term KGs to indicate *directed edge-labeled graphs*.

## 2.1 Knowledge graphs of COVID-19 related articles

Several studies have constructed KGs of causal networks by extracting knowledge from the research articles related to COVID-19 [11]. Kaggle hosts the COVID-19 Open Research Dataset Challenge (CORD-19) [12], providing a dataset of over 400,000 scholarly articles. Some studies have developed KGs based on this dataset and have linked them to external resources [13, 14].

## 2.2 Biomedical ontologies for COVID-19

From a biomedical perspective, Yousefianzadeh et al. [15] explored ontologies in the field of COVID-19 published in a medical ontology database called BioPortal [16]. Coronavirus Infectious Disease Ontology (CIDO) has been developed as a community-based ontology that supports coronavirus disease knowledge and data standardization, integration, sharing, and analysis [17]. The Infectious Disease Ontology (IDO) has been used to build extensions, focusing on COVID-19 [18]. They have been developed to integrate heterogeneous biomedical information, whereas our ontology targets public health operations. It would be beneficial to bridge the ontologies to infer infection risks based on biomedical evidence (IDO) and public health data (CIRO); however, it is beyond the scope of the present study.

## 2.3 Knowledge graphs and ontologies on public health

Several studies that have applied semantic technology to public health issues.

A method for describing the mobility information of infected patients has been proposed, as Patient Location Ontology-based Data (PLOD) [19]. The study used RDF to express location information of cases released by local governments to alert residents in the regions. PLOD can provide personalized exposure notification to cell-phone users, by combining mobility information of their devices stored by cell phone carriers without violating the privacy of patients and residents [20].

Next, a KG that describes relationships between COVID-19 patients and spatiotemporal information on their behavior has been proposed [21]. Using this KG, an experimental system has been developed to calculate the node similarity based on semantic relationships, filter out superspreaders, according to network centrality, and visualize patients' behavior trajectories. Our KG also shares the same feature of preserving spatiotemporal information through an event-centric schema but differs in that it allows inference of infection risk based on the axioms of ontology.

Finally, COviD-19 Ontology for cases and patient information (CODO) [22] has been proposed as an ontology for publishing KGs of COVID-19 epidemiological data based on FAIR data principles [23]. It also provides examples of SPARQL queries with OWL reasoning, and shows that the CODO can be used to investigate infection routes. The CODO has been developed using agile methods, and the ontology continues to be extended. Additionally, González-Eras et al. [24] claimed the need for integration and interoperability of the various ontologies related to COVID-19. Using ontological engineering processes, they merged CODO and WHO COVID-19 Rapid Version CRF semantic data model (COVIDCRFRAPID) [25].

COVID-19 Diagnosis Ontology (CDO) [26] is an ontology containing diagnosis rules extracted from Chinese government documents. Furthermore, using the Semantic Web Rule Language (SWRL) [27] enables the diagnosis of the suspected cases of infection.

The CODO focuses on inference using human relationships, such as relatives and roommates of the infected person, to infer who needs to be screened. Meanwhile, our CIRO aims to understand the behavior and spatial situation of infected persons. The CDO aims to diagnose

COVID-19 and infers suspected cases of infection based on travel history, contact history, clinical symptoms, and serological examination results from travel to Wuhan and other risk areas.

Therefore, they are useful in the diagnostic process after sufficient contact tracing and detailed data have been obtained. Conversely, the CIRO can infer the risk of COVID-19 infection from the contextual information of an individual's behavior and location. Therefore, CIRO has an advantage and can infer the infection risk at an early phase before serological examination or manual contact tracing. Moreover, it can be used to screen individual who have had close contact.

## 3 COVID-19 infection risk ontology

### 3.1 Ontology modeling

Contact tracing of close contacts has been manually conducted by disease intervention specialists, which has been a bottleneck in the identification of infection routes. Therefore, a flexible spatiotemporal search, using time, place, person, and other arbitrary items as keys, would greatly reduce the workload, by narrowing down the contacts that need to be traced further, on a priority basis. To this end, risks associated with such factors must be computable, and we reused the guidelines formulated by the Japanese government, known as the "5 situations."

The "5 situations" were high-risk situations where transmission of COVID-19 becomes possible: Social gatherings with drinking alcohol, Long feasts in large groups, Conversation without masks, Living together in a small limited space, and Switching locations (for short breaks and for end of daily works). They are formulated from epidemiological surveys and selected for nonexperts to understand their risks. However, the infection risk in individual cases is not limited to the situations, and other factors must be considered for a more accurate risk assessment. Through formal expression of the knowledge of risk factors for each situation, automated estimation of infection risks for individual cases would become possible. However, in the 5 situations, the factors are not organized for automated computation, such as the properties of the place, person's behavior, or temporary situation. Therefore, in this study, we developed (CIRO) [28] by organizing the factors necessary for evaluating infection risk with reference to the 5 situations. The CIRO was constructed in Web Ontology Language (OWL) format based on the following design principles:

- To enable searches using any item as a key, such as location, time, and person.
- To enable inference of behaviors of the person from location information, by compiling location-induced infection risk behaviors.
- To enable inference to the degree of the Three Cs for individual behavioral cases.

Figs 1 and 2 illustrate the class hierarchy of the constructed ontology and the relationship between the main classes, respectively. The main classes of the CIRO are `Event`, `Person`, `Time`, `Place`, `Situation`, `Action`, and `Context`, and their relationships between them are defined. Thus, the data generated based on this class relationship allows the user to retrieve a list of instances of each class and search for their relationships. The main extended classes and their axioms are described as follows.

**3.1.1 Event.** In CIRO, we adopted an event-centric schema that has been extended in its own way, with reference to Event Ontology [29] and SimpleEventModel [30]. The vocabulary from schema.org [31] was reused. An event is an entity that describes "who, when, where, and what" information obtained from the patients' behavior surveys. Therefore, an instance of the `Event` class has `agent`, `location`, `action`, and `time` as ObjectProperties.

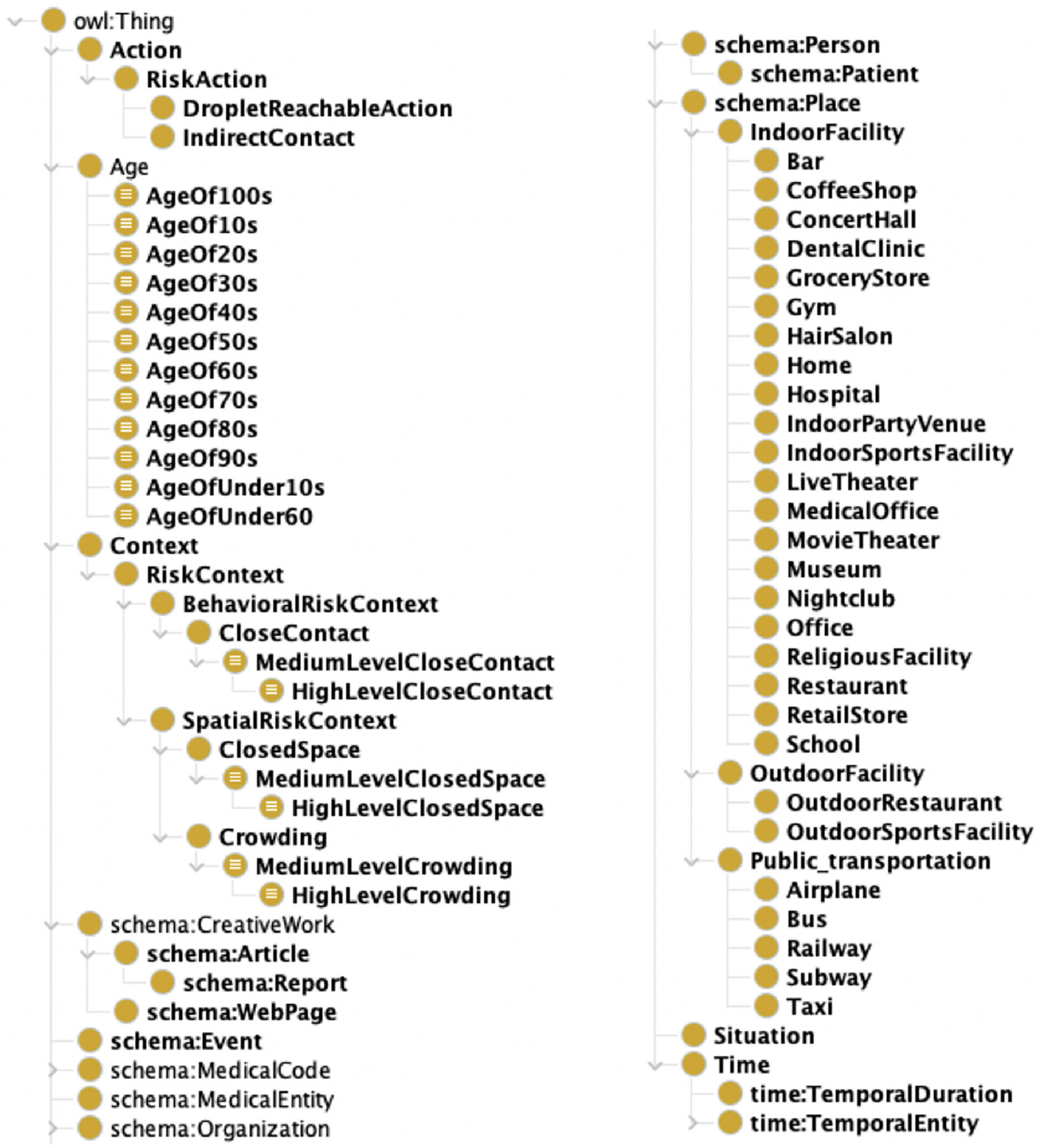

**Fig 1. Part of the CIRO class hierarchy.**

https://doi.org/10.1371/journal.pone.0282291.g001

Additionally, in CIRO, an instance of the `Event` class has a context as described in Section 3.1. It also defines the `followingEvent` property, which describes the relationship before and after the event. At times, the absolute time is unknown during the patients' behavior surveys. In such cases, the pre- and post-event relationships are important alternative information.

**3.1.2 Person.** A person is described as an instance of the `Person` or `Patient` class. In this study, we focused on the spatial and behavioral context to infer the degree of the Three Cs. Once the patient data are accumulated, they can be released as open data, for various analyses by spatial epidemiology after appropriate anoymization. Thus, we simplified the description of health status, residential area, and age groups, in the design, to ease the further processing.

Health status and residential area can be represented by reusing `schema:healthCondition` and `schema:homeLocation`. In CIRO, we defined the `Age` class and further defined the age group classes as its subclasses. The class axioms of each age group

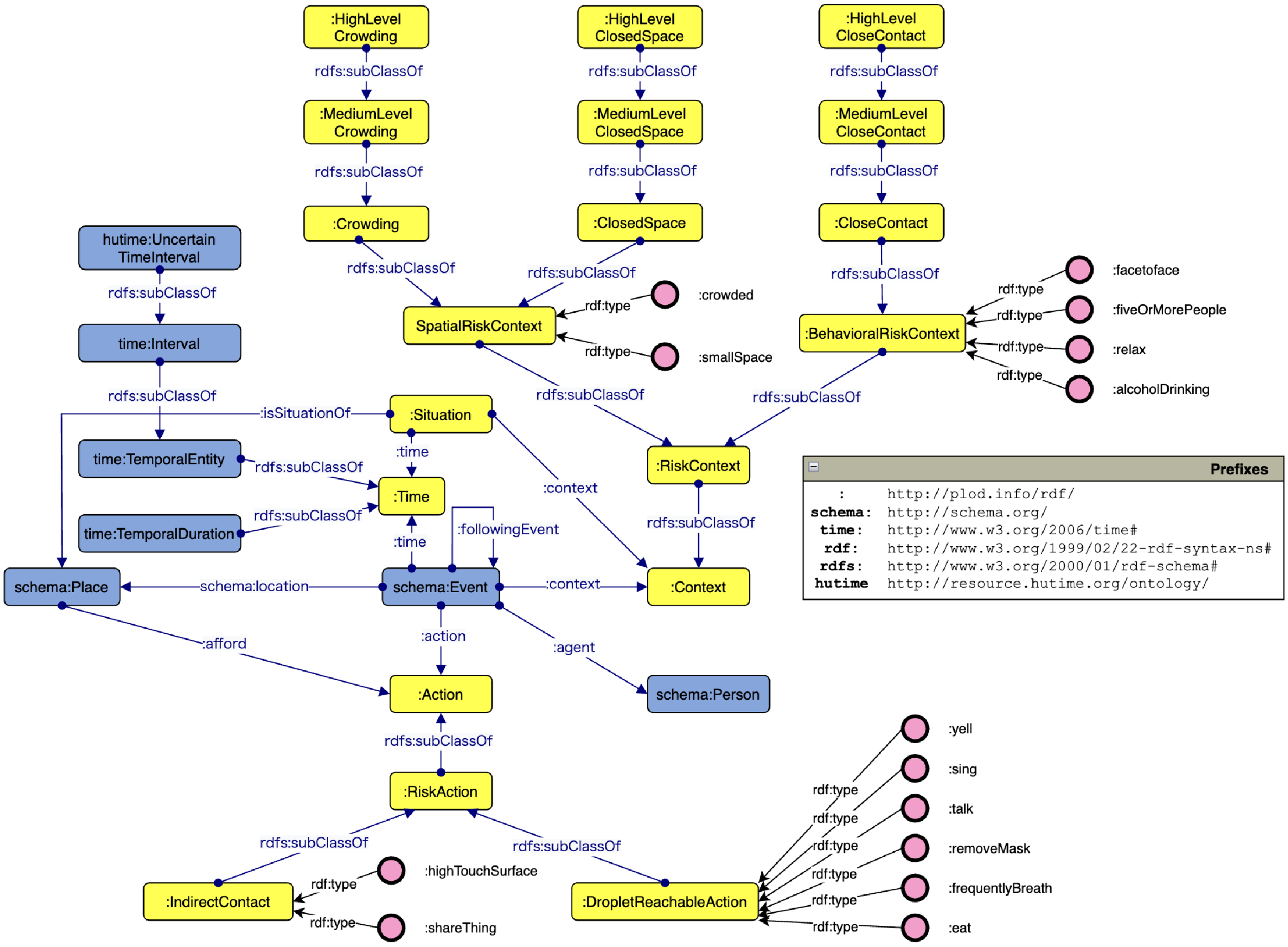

**Fig 2. CIRO class relationship diagram.** Yellow boxes are newly created in this paper, and blue boxes are reused from existing ontologies.

were set as described in Eq 1, representing the 30s.

$$AgeOf30s \equiv Age \sqcap \forall\ value \leq_{30,39} \tag{1}$$

In anonymized RDF data, instances of the `Person` class are connected to such age group classes. When it is necessary to define a new age group for epidemiological analysis, such as 60 years old or younger, the class is defined as in Eq 2.

$$AgeOfUnder60s \equiv Age \sqcap \forall\ value <_{60} \tag{2}$$

Thus, it is possible to flexibly add and classify the necessary age classes after data collection, even with age-anonymized data.

**3.1.3 Time.** Behavior surveys may not reveal an event's start and end times in exact absolute times but only in ambiguous times. Therefore, the CIRO reused the Time Ontology [32] and HuTime [33] to support various time expressions.

To this end, we defined `time:TemporalEntity` and `time:TemporalDuration` as subclasses of the `Time` class. Thus, all values such as temporal instants, intervals, and duration can be used as a temporal representation for `Event` and `Situation`. The `UncertainTimeInterval` class in the HuTime was reused to allow ambiguous time representation. Specifically, an instance of this class has `hasReliableBeginning` to specify a certain starting point of the interval and `hasPossibleJdBeginning` to specify a possible starting point. Thus, it is possible to represent a range of start and end times for the time interval.

We also defined our own `PartOfDay` class and its subclasses as `Morning`, `Afternoon`, `Evening`, and `Night`. These classes were used when only qualitative time information can be obtained during a behavior survey.

**3.1.4 Action.** Various human actions are classified as instances of the `Action` class. The CIRO defines the `RiskAction` class as a subclass of `Action`, and `IndirectContact` and `DropletReachableAction` as its subclasses. The `IndirectContact` is a class of actions that cause indirect contact, and `DropletReachableAction` is a class of actions that may deliver airborne droplets to people nearby. Furthermore, behaviors considered as infection risk factors were extracted from the existing risk indices and then predefined as `NamedIndividuals` of the `IndirectContact` or `DropletReachableAction` classes. Whether or not these risk behaviors occurred during the event is essential information for infection risk inference and epidemiological studies. Additionally, the vocabularies of these behaviors need to be controlled in advance. The relationship between instances of the `Event` class is as follows.

```
@prefix : <http://plod.info/rdf/> .
<http://plod.info/rdf/id/event_0>
  a schema:Event ;
  :action :talk .
```

In future, if the epidemiology of COVID-19 progresses and new behaviors of concern are identified, the CIRO can be extended by adding new `RiskAction` subclasses and `NamedIndividuals`.

**3.1.5 Place.** We defined the subclasses of `schema:Place` such as indoor facilities, outdoor facilities, and public transportation. We also defined various classes of facilities in those subclasses. Furthermore, we introduced affordance to describe the relationship between places and actions. For example, the relationship between "what kind of infection risk action a place, such as a store, facility, or public transportation potentially enables for humans" is defined as a class axiom. For example, the relationship "Restaurant" affords the action of "remove mask" is defined as the following description logic Eq 3 as a class axiom for the `Restaurant` class.

$$Restaurant \sqsubseteq \exists\, afford.\{removeMask\} \tag{3}$$

The axioms of Eqs 4 and 5 are also added because restaurants induce conversation and sharing of objects.

$$Restaurant \sqsubseteq \exists\, afford.\{talk\} \tag{4}$$

$$Gym \sqsubseteq \exists\, afford.\{shareThing\} \tag{5}$$

There may be cases where it is extremely difficult to obtain information about the patient's risk action at the event during the patient's behavior survey. It is possible to infer the potential risk of the event in such cases, by defining the affordance relationship between place and risk action.

**3.1.6 Situation.** CIRO defines the `Situation` as a class that represents the spatial situation of a certain place at a certain time, where an instance $e_i \in Event$ of the `Event` class has the `schema:location` property, and its value is a subclass when an instance $p_i \in Place$ of the Situation class exists, an instance $s_i \in Situation$ of the Situation class also exists, and the value of its `isSituationOf` property is $p_i$. The `Situation` class is separated from the `Event` class, which represents behavioral information because the `Situation` class represents the spatial situation of a place.

**3.1.7 Context.** In this study, we introduced the concept of *context* to attach additional or contextual semantic information to events and situations. Here, we created a subclass `RiskContext` for specializing in the context of infection risk because `Context` is an abstract class. As subclasses of `RiskContext`, we defined `SpatialRiskContext` to indicate risks caused by spatial events, and `BehavioralRiskContext` to indicate risks caused by behavioral events. Additionally, risk factors that further explain space and behavior were extracted from existing risk indices and predefined as `NamedIndividuals`. For example, the relationship between instances of the `Event` class and instances of the `Situation` class is as follows.

```
@prefix : <http://plod.info/rdf/> .
<http://plod.info/rdf/id/event_0>
  a schema:Event ;
  :action :talk ;
  schema:location <http://plod.info/rdf/id/Bus_0> ;
  :context :facetoface, :relax .

<http://plod.info/rdf/id/situation_0>
  a :Situation ;
  :isSituationOf <http://plod.info/rdf/id/Bus_0> ;
  :context :crowded, :smallSpace .
```

We also defined the classes corresponding to the Three Cs, which comprise `CloseContact`, `ClosedSpace`, and `Crowding`, as a subclass of `SpatialRiskContext` and `BehavioralRiskContext`. The levels of Three Cs were defined as low (no risk), medium (medium closeness, medium crowdedness, and medium enclosedness), and high (high closeness, high crowdedness, and high enclosedness). The medium and high levels excluding no risk were defined as subclasses of each of the Three Cs classes. We also defined the axioms corresponding to these classes. Thus, individual events and situations are classified into these classes by OWL reasoning. The high- and medium-level risk classes were defined as the Eqs 6–11, respectively.

$$
\begin{aligned}
& HighLevelClosedSpace \\
& \quad \equiv \exists\ isSituationOf.(IndoorFacility \\
& \quad \sqcup\ Public_transportation) \\
& \quad \sqcap\ \geq\ 1\ context.SpatialRiskContext
\end{aligned} \tag{6}
$$

$$
\begin{aligned}
& HighLevelCrowding \\
& \quad \equiv\ \geq\ 1\ context.BehavioralRiskContext \\
& \quad \sqcap\ \geq\ 2\ context.SpatialRiskContext
\end{aligned} \tag{7}
$$

$$\begin{aligned} & HighLevelCloseContact \\ & \equiv \exists\ location.(\geq\ 2\ afford.DropletReachableAction) \\ & \sqcup\ \geq\ 2\ action.DropletReachableAction \\ & \sqcap\ \exists\ time.(\exists\ hasDuration.(\exists\ numericDuration.>_{15})) \\ & \sqcap\ \geq\ 1\ context.BehavioralRiskContext \end{aligned} \tag{8}$$

$$\begin{aligned} & MediumLevelClosedSpace \\ & \quad \equiv \exists\ isSituationOf.(IndoorFacility \\ & \quad \sqcup\ Public_transportation) \end{aligned} \tag{9}$$

$$\begin{aligned} & MediumLevelCrowding \\ & \quad \equiv\ \geq\ 1\ context.BehavioralRiskContext \\ & \quad \sqcap\ \geq\ 1\ context.SpatialRiskContext \end{aligned} \tag{10}$$

$$\begin{aligned} & MediumLevelCloseContact \\ & \equiv \exists\ location.(\geq\ 1\ afford.DropletReachableAction) \\ & \sqcup\ \geq\ 1\ action.DropletReachableAction \\ & \sqcap\ \exists\ time.(\exists\ hasDuration.(\exists\ numericDuration.>_{15})) \\ & \sqcap\ \geq\ 1\ context.BehavioralRiskContext \end{aligned} \tag{11}$$

For example, Eq 6 indicates that an entity with at least one of the values of the `isSituationOf` being an indoor facility or public transportation and with one or more `SpatialRiskContext` instances is inferred as an instance of `HighLevelClosedSpace`. Here, the difference between high-level and medium-level risks depends on the conditions of the inference rules, which do not differ in any other qualitative semantics. All the class axioms of medium-level risks are relaxed conditions of high-level risks. Thus, for example, if an event is inferred to be high-level close contacts, it is also inferred to be medium-level close contacts. To achieve such inference, we defined high-level risk as a subclass of medium-level risk.

Thus, for individual cases, it is possible to infer the degree of the Three Cs based on the risk context of the behavior and space at the time. The cardinality of each class axiom of the Three Cs is empirically set in this study. These values can be adjusted based on future epidemiological surveys, and the inference conditions can be flexibly set without altering the obtained data.

## 3.2 Ontology-based knowledge graph construction

A total of 247 action scenarios including locations, times, and situations were entered into a spreadsheet to simulate patients' behavior surveys. They were used as simulated data. These data were converted into a KG in RDF based on the CIRO. We used TogoDB [34] to convert the tabular data to the RDF data. Fig 3 depicts an example of the KG of the event "Person A had dinner with B and C at a restaurant from 12:00 to 13:00 on April 1, 2020." The magenta arrows represent the relationships derived by OWL reasoning. The action of "remove mask" is induced since the location is a restaurant. Therefore, even if the behavior survey did not reveal the information that the mask was removed, it is considered as a risky action that might have occurred in this event. OWL reasoning classifies `Event` and `Situation` instances into Three Cs classes. We generated KG for each behavioral event and stored the data in 6,924 triples in the triple store GraphDB [35].

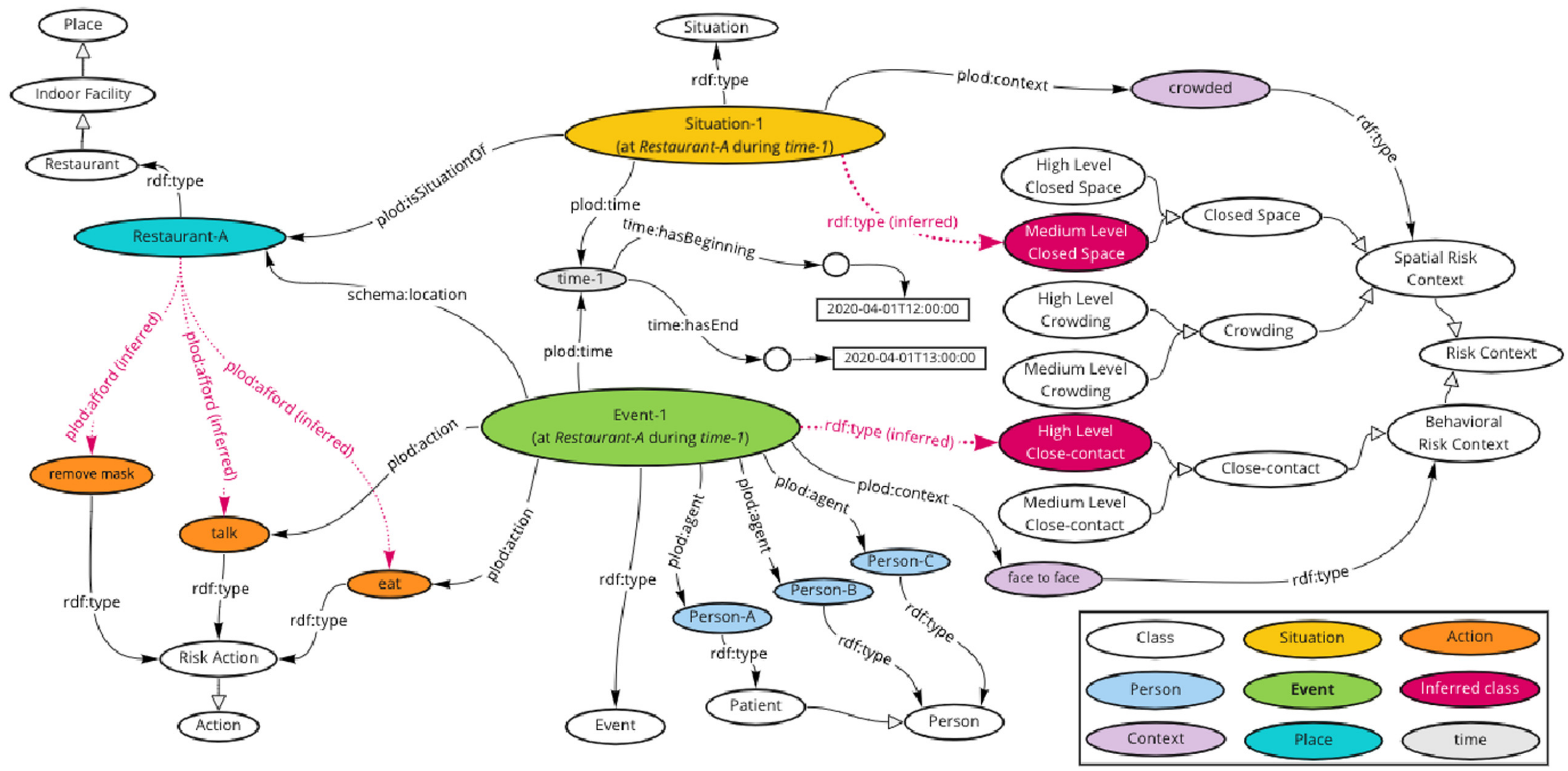

**Fig 3. Example of a constructed knowledge graph.**

### 3.3 Querying knowledge graphs with reasoning

GraphDB enables users to retrieve data from RDF data using SPARQL. For example, it is possible to search for another behavioral history that intersects with any given action history in time and space (Fig 4) to identify infection routes.

GraphDB supports several OWL subsets [36] such as OWL 2 QL and OWL 2 RL in addition to RDFS [37]. Any relations that are not explicitly described in the stored RDF data can be derived based on these inference rules. Therefore, the SPARQL query in Fig 5 can retrieve the persons attending events with the same enclosed level as any given person. The results are indicated in descending order by event count (Fig 6).

Furthermore, even users unfamiliar with SPARQL can intuitively conduct contact tracing of close contacts by visualizing the graph structure of the KG. Fig 7 shows an example of visualizing a KG using the visual graph function of GraphDB. The user can freely expand each node to visually track the location of a cluster and the number of close contacts.

## 4 Evaluation

### 4.1 Ontology consistency

It is essential to evaluate whether the constructed ontology's class and property axioms have consistency in order to query ontology-based data with reasoning. Thus, we used OntoDebug [38], a plugin for Protégé [39], to validate the consistency of CIRO. The OntoDebug is an

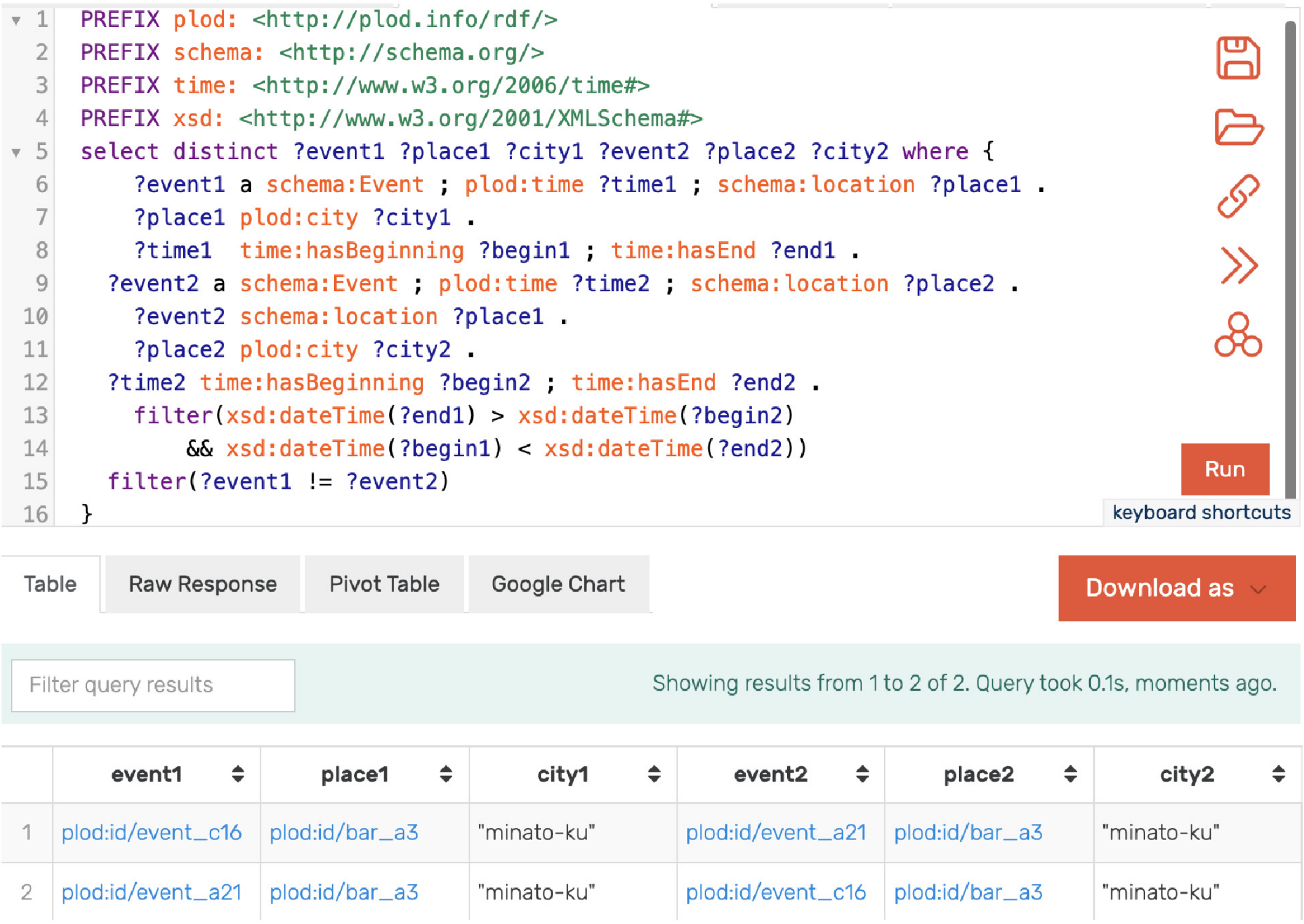

**Fig 4. SPARQL query and its result to find two events that intersect in time and space.**

https://doi.org/10.1371/journal.pone.0282291.g004

interactive ontology debugger that implements many black-box algorithms proposed for ontology debugging over the recent decade. The results of running OntoDebug proved that CIRO is a coherent and consistent ontology.

## 4.2 Reasoning efficiency

In this section, we evaluate the executability and execution time of CIRO-based reasoning for varying data sizes and spatiotemporal conditions.

```
PREFIX plod: <http://plod.info/rdf/>
PREFIX schema: <http://schema.org/>
PREFIX time: <http://www.w3.org/2006/time#>
PREFIX xsd: <http://www.w3.org/2001/XMLSchema#>
PREFIX rdfs: <http://www.w3.org/2000/01/rdf-schema#>
SELECT ?agent (count(distinct ?event) AS ?cnt)  WHERE {
    ?event plod:agent <http://plod.info/rdf/id/person_a_A> ;
        plod:agent ?agent ;
        schema:location ?place .
    filter (?agent != <http://plod.info/rdf/id/person_a_A>)
    ?situation plod:isSituationOf ?place .
    ?situation a plod:ClosedSpace .
} GROUP BY ?agent ORDER BY DESC(?cnt)
```

**Fig 5. SPARQL query to retrieve the people who appear with *Person A* in the event of a closed situation, in descending order by the number of that events.**

https://doi.org/10.1371/journal.pone.0282291.g005

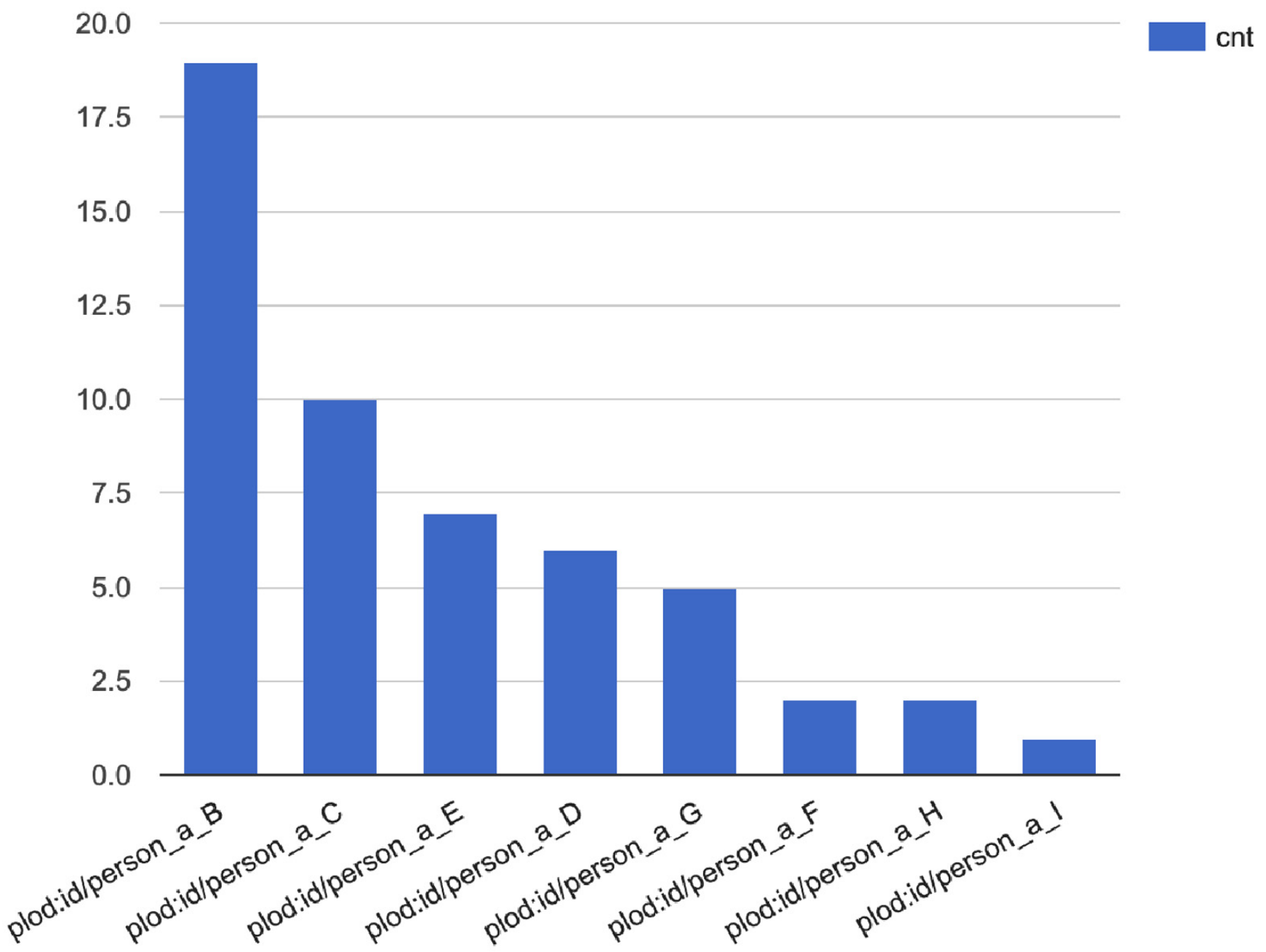

**Fig 6. Visualization results of Fig 5.** The horizontal axis represents the people who appear with *Person A* in the events of a closed situation, and the vertical axis represents the number of its appearance.

We employ axioms that OWL 2 DL [40] can use to infer, such as in Eq 8. However, GraphDB does not support OWL 2 DL inference, making it impossible to verify the validity of inference for all the Three Cs. Therefore, in this study, we use HermiT [41], an inference engine that supports OWL 2 DL, to verify whether it can correctly infer all infection risks using CIRO, and measure the inference time.

The test data (pseudo-action records) used in the inference experiments are RDF data generated based on the CIRO schema with randomly changed values for time, location, action, and situation. The inference is made according to the infection risk that the `Event` and `Situation` instances in this test data are classified as.

Here, infection risk refers to the degree of the Three Cs. It consists of six items: high closeness, high crowdedness, high enclosedness, medium closeness, medium crowdedness, and medium enclosedness. The total number of items becomes nine by adding the low-risk categories of low closeness, low crowdedness, and low enclosedness. An event or a situation may also have multiple infection risks at one time. Therefore, there are $3 \times 3 \times 3 = 27$ combinations of infection risks. The proportions of high risk are 10%, 20%, and 30%, respectively, and those of medium risk are 20%, 40%, and 60%, respectively, generating $27 \times 9 = 243$ test datasets. The number of test data was set to three patterns: 100, 500, and 1000. For example, in the test case one, the risk of closeness is varied from "high," "medium," and "low (no risk)," and the risks of crowdedness and enclosedness are fixed at "high" to generate data. Out of 100 data, the percentages of events satisfying simultaneously high closeness, high crowdedness, and high enclosedness; medium closeness, high crowdedness, and high enclosedness; and low closeness, high crowdedness, and high enclosedness are 10%, 20%, and 70%, respectively. Therefore, the test data are generated so that the number of correct data (events) for the Three Cs in the test case one are: 10 high-closeness, 100 high-enclosedness, 100 high-crowdedness, 20 medium-closeness, 0 medium-enclosedness, 0 medium-crowdedness, 70 low-closeness, 0 low-enclosedness, and 0 low-crowdedness.

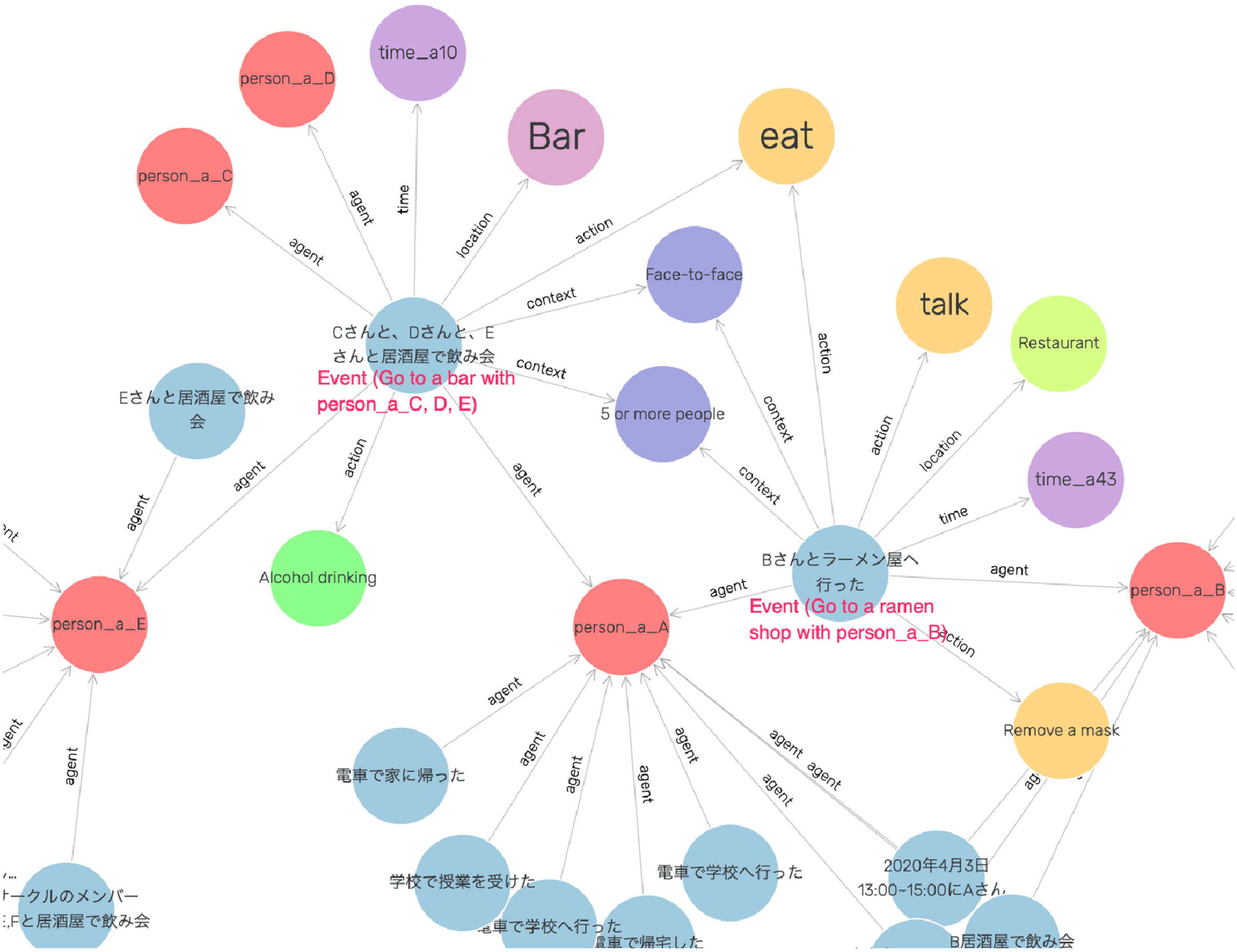

**Fig 7. Examples of knowledge graph visualization.**

https://doi.org/10.1371/journal.pone.0282291.g007

In this experiment, we used Owlready2 [42], a Python library that can use the inference engine HermiT. We issued SPARQL queries to obtain instances of the Three Cs classes after performing inference with HermiT. We then counted the number of inferred instances and measured processing times. The machine used for the experiments was an AMD EPYC 7402P 24-Core Processor CPU with 500 Gbytes of main memory and CentOS7 OS.

We confirmed that the inference was valid in the all test cases. Table 1 presents the total number of the data and inference time. We also published the data, inference results, and experimental scripts for all test cases [43]. The results show that there is a large difference between the minimum and maximum inference time. The test cases that require more inference time contain more high closeness and high crowdedness events, wherease those with shorter inference time do not contain these events. The class axioms for high closeness and high crowdedness include the condition “the cardinality is greater than or equal to 2,” which is thought to affect inference time significantly.

Table 1. Number of data and inference time (in seconds).

| Data size | Average | Min | Max | Median |
|---|---|---|---|---|
| 100 | 2.95 | 0.91 | 9.76 | 2.06 |
| 500 | 76.82 | 2.17 | 399.63 | 32.75 |
| 1,000 | 417.70 | 5.33 | 2,110.31 | 215.00 |

https://doi.org/10.1371/journal.pone.0282291.t001

# 5 Discussion

## 5.1 Ontological perspective

To maximize the benefits of the CIRO in active epidemiological surveillance, we discuss the sufficiency and extensibility of the ontology. As stated in Sections 3.1 and 4.2, we have confirmed that the CIRO is useful in searching for close contacts by using arbitrary items, such as location and time. Moreover, it can also be used to infer the degree of the Three Cs. The introduction of this ontology in public health operations would reduce human interventions in contact tracing.

This study defines risky instances of behaviors and spatial context based on indicators identified by past statistical analysis. In the future, more factors might be considered as risks, by confirming new statistical facts, or by emerging variants that have different transmission patterns. CIRO can be extended to add new risk factors by defining them as instances of `RiskContext` or `RiskAction` subclasses.

In this study, we tentatively defined the class axioms for the Three Cs (Eq 6–11); however, different inferences can be made by changing the conditions within the limitations of the OWL 2 DL. In short, CIRO can flexibly extend the definitions of COVID-19 infection risks even when new findings contradict the past findings. Moreover, we believe that it will be possible to identify which behaviors and spatial conditions are inherently at infection risks by conducting inference experiments with incremental changes in the axioms and comparing them with data from positive individuals.

Thus, CIRO is extensible enough to catch up with the changes in infection risks of COVID-19. Furthermore, excluding these COVID-19-specific classes and instances, it becomes a general ontology for risky events and situations. Thus, by reusing the core model of CIRO and modifying the subclasses and instances, there is a possibility that the model can be applied to the knowledge representation of infection risks of infectious diseases other than COVID-19.

## 5.2 Practical perspective

There are several issues to be resolved before CIRO can be integrated into practical systems. In developing and assessing the ontology, we have assumed a system for infection risk control by regional public health centers (Fig 8), which works as follows.

In this framework, patients report their behaviors to the behavior reporting system, owned by the public health authority, to identify the infection route (contact tracing) and to alert their potential contacts about the infection risk. Then, public health officials interview the patient by referencing the information stored in the system, such as visiting places, events, and their actions. In this process, CIRO-based reasoning would efficiently guide the patient interviews. If there is any visits or actions related to infection risk, the public officials update the patient data. For example, if the patient were not wearing a mask during a meeting, it would be labeled as such.

As the information on the patients' behavior accumulates, in this manner, automated assessment of infection risks would become possible, which have been unknown neither by

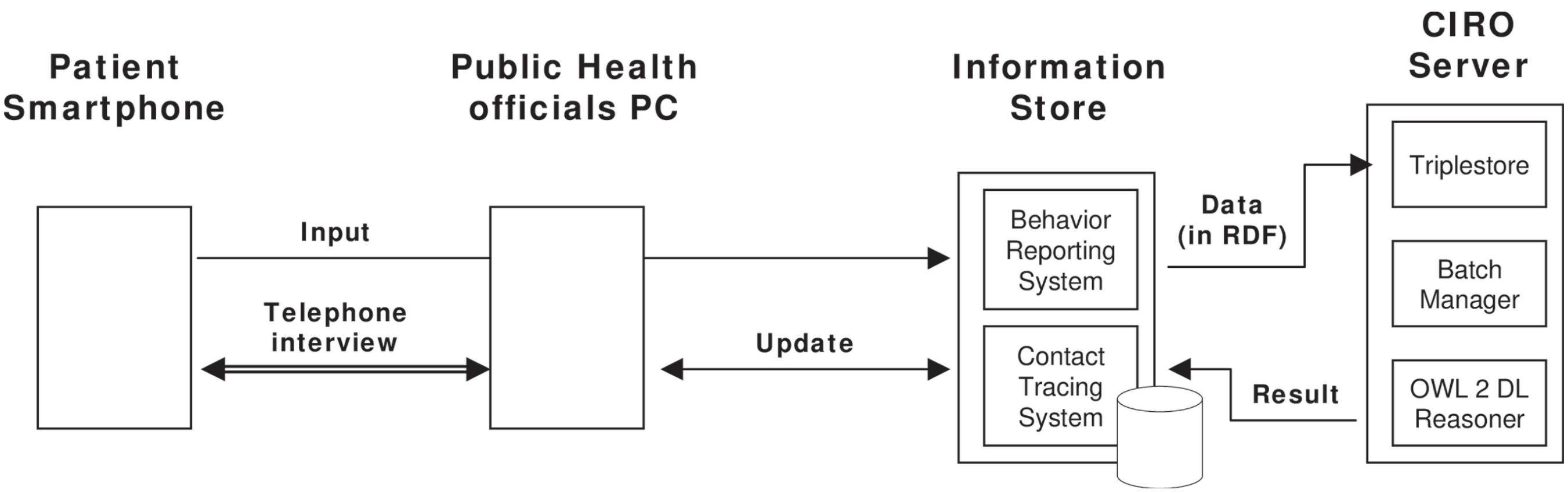

**Fig 8. Behavior reporting and Infection risk assessment.**

https://doi.org/10.1371/journal.pone.0282291.g008

patients nor public health officials. For example, intersections with infected individuals could be found by identifying their overlaps at a certain time in a public place.

However, the processing time increases as the number of data increases, making it challenging to obtain results in a realistic amount of time. Preliminary experiments with 10,000 data points showed that results could only be obtained after up to 10 h, suggesting that real-time SPARQL queries with OWL reasoning based on CIRO require data distribution and parallel processing. Alternatively, CIRO could only use simpler expressions such as OWL 2 EL, QL, and RL, to increase the inference efficiency. Specifically, instead of using a class expression with cardinality, such as "two or more," it specifies instances directly with `hasValue` and enumerates combinations of two instances. However, this method causes a combinatorial explosion when the number of instances increases, making it challenging to manage the rules.

These factors must be verified to be a practical solution. Accordingly, we are collaborating with the prefectural government and setting up feasibility studies to prove the utility of our proposal.

## 6 Conclusions

In this study, we constructed an ontology, CIRO, which can infer the risk of COVID-19 infection for the actual operation of tracking and screening of close contacts at public health centers. For evaluation, we built KG using dummy patients' information, and confirmed that the ontology enabled flexible searches by various keys, such as location and time. We also evaluated the computational efficiency of the KG.

The main contributions of this study are summarized as follows: First, we developed a CIRO that can organize the risk of COVID-19 infection related to place and behavior and infer the degree of infection risk. Second, we demonstrated a method to transform action history into KGs based on the CIRO for useful graph retrieval for contact tracing. Finally, we evaluated the ontology qualitatively and quantitatively, clarifying the considerations needed for practical implementation.

We are developing end-user applications that can be used by disease intervention specialists for active epidemiological surveillance and setting up field studies to evaluate and revise the CIRO into more practical forms. We envisage that CIRO would be used in public health centers, by integrated within systems to organize information about patients and their potential contacts. Furthermore, the actual patient data would accumulate and lead to the identification of new risk factors automatically through searches and inference on the KG technologies.

## Acknowledgments

We thank Mr. Masahide Kanzaki (Zenon Limited Partners), Mr. Shoichi Sakane (Cisco Systems G.K.), Ms. Maori Ito (PLOD info), and Ms. Masako Nomoto (RIKEN) for their various assistance in conducting this research. We would like to express our deepest gratitude.

## Author Contributions

**Conceptualization:** Shusaku Egami, Yasunori Yamamoto, Ikki Ohmukai.

**Data curation:** Shusaku Egami, Yasunori Yamamoto, Ikki Ohmukai.

**Formal analysis:** Shusaku Egami, Yasunori Yamamoto, Ikki Ohmukai.

**Funding acquisition:** Takashi Okumura.

**Investigation:** Shusaku Egami, Ikki Ohmukai.

**Project administration:** Takashi Okumura.

**Resources:** Yasunori Yamamoto.

**Software:** Shusaku Egami.

**Supervision:** Yasunori Yamamoto, Ikki Ohmukai.

**Writing – original draft:** Shusaku Egami.

**Writing – review & editing:** Yasunori Yamamoto, Ikki Ohmukai, Takashi Okumura.

## References

1. MacDonald PD. Methods in field epidemiology. Jones & Bartlett Publishers; 2011.
2. Chan LYH, Yuan B, Convertino M. COVID-19 non-pharmaceutical intervention portfolio effectiveness and risk communication predominance. Scientific reports. 2021; 11(1):1–17. https://doi.org/10.1038/s41598-021-88309-1 PMID: 34012040
3. Emanuel EJ, Philips JP, Popescu S. COVID-19 Activity Risk Index; 2020. https://www.covid19reopen.com/resources/covid-19-daily-activity-risk-index.
4. Prime Minister's Office of Japan. Avoid the "Three Cs"!; 2020. https://www.mhlw.go.jp/content/3CS.pdf.
5. RDF Core Working Group. RDF; 2014. https://www.w3.org/RDF/.
6. Harris S, Seaborne A. SPARQL 1.1 Query Language; 2013. https://www.w3.org/TR/sparql11-query/.
7. OWL Working Group. OWL; 2012. https://www.w3.org/OWL/.
8. Hogan A, Blomqvist E, Cochez M, D'amato C, Melo GD, Gutierrez C, et al. Knowledge Graphs. ACM Comput Surv. 2021; 54(4). https://doi.org/10.1145/3447772
9. Ehrlinger L, Wöß W. A knowledge graph acquires and integrates information into an ontology and applies a reasoner to derive new knowledge. Towards a Definition of Knowledge Graphs. SEMANTiCS (Posters, Demos, SuCCESS). 2016; 48(1-4):2.
10. Feilmayr C, Wöß W. An analysis of ontologies and their success factors for application to business. Data & Knowledge Engineering. 2016; 101:1–23. https://doi.org/10.1016/j.datak.2015.11.003
11. Domingo-Fernández D, Baksi S, Schultz B, Gadiya Y, Karki R, Raschka T, et al. COVID-19 Knowledge Graph: a computable, multi-modal, cause-and-effect knowledge model of COVID-19 pathophysiology. Bioinformatics. 2020; 37(9):1332–1334. https://doi.org/10.1093/bioinformatics/btaa834
12. Allen Institute for AI. COVID-19 Open Research Dataset Challenge (CORD-19); 2019. https://www.kaggle.com/datasets/allen-institute-for-ai/CORD-19-research-challenge.

13. Michel F, Gandon F, Ah-Kane V, Bobasheva A, Cabrio E, Corby O, et al. Covid-on-the-Web: Knowledge Graph and Services to Advance COVID-19 Research. In: International Semantic Web Conference; 2020. p. 294–310.
14. Steenwinckel B, Vandewiele G, Rausch I, Heyvaert P, Taelman R, Colpaert P, et al. Facilitating the Analysis of COVID-19 Literature Through a Knowledge Graph. In: International Semantic Web Conference; 2020. p. 344–357.
15. yousefianzadeh o, Taheri A. COVID-19 ontologies and their application in medical sciences: Reviewing Bioportal. Applied Health Information Technology. 2020; 1(1):30–35.
16. National Center for Biomedical Ontology. BioPortal; 2005. https://bioportal.bioontology.org/.
17. He Y, Yu H, Ong E, Wang Y, Liu Y, Huffman A, et al. CIDO, a community-based ontology for coronavirus disease knowledge and data integration, sharing, and analysis. Scientific Data. 2020; 7(1):1–5. https://doi.org/10.1038/s41597-020-0523-6 PMID: 32533075
18. Babcock S, Beverley J, Cowell LG, Smith B. The infectious disease ontology in the age of COVID-19. Journal of biomedical semantics. 2021; 12(1):1–20. https://doi.org/10.1186/s13326-021-00245-1 PMID: 34275487
19. Ohmukai I, Yamamoto Y, Ito M, Okumura T. Tracing patient PLOD by mobile phones Mitigation of epidemic risks based on patient locational open data. In: 2020 IEEE 29th International Conference on Enabling Technologies: Infrastructure for Collaborative Enterprises (WETICE); 2020. p. 283–286.
20. Ami J, Ishii K, Sekimoto Y, Masui H, Ohmukai I, Yamamoto Y, et al. Computation of Infection Risk via Confidential Locational Entries: A Precedent Approach for Contact Tracing With Privacy Protection. Ieee Access. 2021; 9:87420–87433. https://doi.org/10.1109/ACCESS.2021.3087478
21. Jiang B, You X, Li K, Li T, Zhou X, Tan L. Interactive Analysis of Epidemic Situations Based on a Spatiotemporal Information Knowledge Graph of COVID-19. IEEE Access. 2020;Early Access:1–14. https://doi.org/10.1109/ACCESS.2020.3033997 PMID: 35937640
22. Dutta B, DeBellis M. CODO: An Ontology for Collection and Analysis of COVID-19 Data. In: Proceedings of the 12th International Joint Conference on Knowledge Discovery, Knowledge Engineering and Knowledge Management. INSTICC. SciTePress; 2020. p. 76–85.
23. GO FAIR. FAIR Principles; 2016. https://www.go-fair.org/fair-principles/.
24. González-Eras A, Santos RD, Aguilar J, Lopez A. Ontological engineering for the definition of a COVID-19 pandemic ontology. Informatics in Medicine Unlocked. 2022; 28:100816. https://doi.org/10.1016/j.imu.2021.100816 PMID: 34934805
25. Luiz Bonino. WHO COVID-19 Rapid Version CRF semantic data model; 2020. https://bioportal.bioontology.org/ontologies/COVIDCRFRAPID.
26. Wu H, Zhong Y, Tian Y, Jiang S, Luo L. Automatic diagnosis of COVID-19 infection based on ontology reasoning. BMC medical informatics and decision making. 2021; 21(9):1–13. https://doi.org/10.1186/s12911-021-01629-0 PMID: 34789243
27. Horrocks I, Patel-Schneider PF, Boley H, Tabet S, Grosof B, Dean M, et al. SWRL: A semantic web rule language combining OWL and RuleML. W3C Member submission. 2004; 21(79):1–31.
28. Egami S, Yamamoto Y, Ohmukai I, Okumura T. CIRO; 2022. Available from: https://github.com/PLOD-info/PLOD/blob/master/rdf/CIRO.owl.
29. Raimond Y, Abdallah S. The Event Ontology; 2007. http://motools.sourceforge.net/event/event.html.
30. Van Hage WR, Malaisé V, Segers R, Hollink L, Schreiber G. Design and use of the Simple Event Model (SEM). Journal of Web Semantics. 2011; 9(2):128–136. https://doi.org/10.1016/j.websem.2011.03.003
31. Guha RV, Brickley D, Macbeth S. Schema.org: evolution of structured data on the web. Communications of the ACM. 2016; 59(2):44–51. https://doi.org/10.1145/2844544
32. Cox S, Little C. Time Ontology in OWL. W3C Candidate Recommendation Draft, W3C. 2022;.
33. Sekino T. Data description and retrieval using periods represented by uncertain time intervals. Journal of Information Processing. 2020; 28:91–99. https://doi.org/10.2197/ipsjjip.28.91
34. Database Center for Life Science. DBCLS TogoDB database hosting service; 2008. http://togodb.org/.
35. Ontotext USA, Inc. Ontotext GraphDB; 2022. https://www.ontotext.com/products/graphdb/.
36. Motik B, Cuenca Grau B, Horrocks I, Wu Z, Fokoue A, Lutz C. OWL 2 web ontology language: Structural specification and functional-style syntax. W3C recommendation. 2012;.
37. Brickley D, Guha RV, McBride B. RDF Schema 1.1. W3C recommendation. 2014; 25:2004–2014.
38. Schekotihin K, Rodler P, Schmid W. OntoDebug: Interactive Ontology Debugging Plug-in for Protégé. In: Ferrarotti F, Woltran S, editors. Foundations of Information and Knowledge Systems. Springer International Publishing; 2018. p. 340–359.

39. Stanford University. Protégé; 2016. https://protege.stanford.edu/.
40. Motik B, Patel-Schneider PF, Parsia B, Bock C, Fokoue A, Haase P, et al. OWL 2 web ontology language: Structural specification and functional-style syntax. W3C recommendation. 2009; 27(65):159.
41. Glimm B, Horrocks I, Motik B, Stoilos G, Wang Z. HermiT: an OWL 2 reasoner. Journal of Automated Reasoning. 2014; 53(3):245–269. https://doi.org/10.1007/s10817-014-9305-1
42. Lamy JB. Owlready: Ontology-oriented programming in Python with automatic classification and high level constructs for biomedical ontologies. Artificial intelligence in medicine. 2017; 80:11–28. https://doi.org/10.1016/j.artmed.2017.07.002 PMID: 28818520
43. Egami S, Yamamoto Y, Ohmukai I, Okumura T. CIRO experimental results; 2022. Available from: https://doi.org/10.5281/zenodo.6482275.